\documentclass[a4paper,conference]{IEEEtran}
\usepackage{cite}
\usepackage{amsmath,amssymb,amsfonts}
\usepackage{algorithmic}
\usepackage{graphicx}
\usepackage{textcomp}
\usepackage{xcolor}
\usepackage{microtype}
\usepackage{booktabs}
\usepackage{tabularx}

\usepackage{pifont}
\usepackage{multirow}
\usepackage{multicol}

\def\BibTeX{{\rm B\kern-.05em{\sc i\kern-.025em b}\kern-.08em
    T\kern-.1667em\lower.7ex\hbox{E}\kern-.125emX}}
\newcommand{\cmark}{\ding{51}}%
\newcommand{\xmark}{\ding{55}}%
\begin{document}

\title{FastSal: a Computationally Efficient Network for Visual Saliency Prediction}

\author{\IEEEauthorblockN{Feiyan Hu and Kevin McGuinness}
\IEEEauthorblockA{Insight Centre for Data Analytics, Dublin City University\\\small
\{feiyan.hu, kevin.mcguinness\}@dcu.ie}
}

\maketitle

\begin{abstract}
This paper focuses on the problem of visual saliency prediction, predicting regions of an image that tend to attract human visual attention, under a constrained computational budget. We modify and test various recent efficient convolutional neural network architectures like EfficientNet and MobileNetV2 and compare them with existing state-of-the-art saliency models such as SalGAN and DeepGaze II both in terms of standard accuracy metrics like AUC and NSS, and in terms of the computational complexity and model size. We find that MobileNetV2 makes an excellent backbone for a visual saliency model and can be effective even without a complex decoder. We also show that knowledge transfer from a more computationally expensive model like DeepGaze II can be achieved via pseudo-labelling an unlabelled dataset, and that this approach gives result on-par with many state-of-the-art algorithms with a fraction of the computational cost and model size. 
\end{abstract}


\section{Introduction}

Eye tracking experiments have shown that human beings are remarkably consistent in where they attend to when exposed to a visual stimulus. The attended areas, known as salient regions, usually contain important information and are analysed by the visual cortex at a higher resolution through directed foveation. Indeed, large changes in visual content outside the salient regions can often go unnoticed, a phenomena known as inattentional blindness~\cite{mack1998inattentional}.

Efforts to create effective models of human visual attention have made rapid progress since the seminal work of Itti et al.~\cite{itti1998model}. Early approaches, largely based on engineered features and heuristics, have been replaced by more accurate and more complex deep neural models (e.g.~\cite{pan2016shallow,pan2017salgan,kruthiventi2017deepfix,kummerer2016deepgaze,jia2020eml,cornia2018predicting,fan2018emotional}). This shift was made possible by the release of various large annotated datasets. Crowdsourced datasets like iSUN~\cite{xu2015turkergaze}, which uses widely available but less accurate webcam-based eyetracking, and SALICON~\cite{jiang2015salicon}, which is based on artificial foevation and needs only a computer with a mouse, are much larger than their predecessors and have allowed the training of deep models with millions of parameters. 


Effective computational visual attention models have many potential applications. For example, Mohedano et al.~\cite{mohedano2018saliency} use visual saliency to improve the performance of image retrieval engines. Gurum Munirathnam et al.~\cite{venkatesh2019saliency} recently showed that saliency can be used to provide cues to improve object detection. Saliency models have also been used in video survellience systems~\cite{yubing2011spatiotemporal}, video compression~\cite{hadizadeh2013saliency}, image retargetting~\cite{chen2003visual}, and various other applications~\cite{mancas2016applications}.


One of the most promising applications of computational visual attention models is in their potential to reduce the complexity of scene analysis and speed up downstream tasks. Indeed, Itti et al.~\cite{itti1998model, itti2001computational} list this as one of the key motivations in their seminal work on computational visual attention. Unfortunately, many existing deep saliency models are as computationally demanding, if not more so, than the subsequent visual analysis steps (classification, object detection, retrieval, etc.).  This has limited their adoption in downstream tasks.

\begin{figure}
    \centering
    \includegraphics[width=1.0\columnwidth]{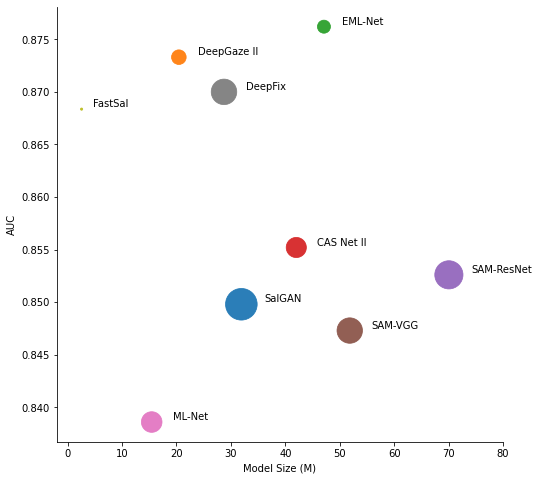}
    \caption{Model size vs AUC on MIT 300 for various popular image saliency models. Larger dots represent more computationally complex models (see Table~\ref{tab:flops_size}). }
    \label{fig:my_label}
\end{figure}


This paper focuses on computationally efficient approaches to visual saliency, aiming to retain near state-of-the-art performance at a small fraction of the computational cost. We benchmark many existing deep saliency models in terms of accuracy metrics (AUC, CC, NSS, etc.), theoretical complexity (GFLOPs and model size), and actual computational performance on both CPU and GPU hardware. We further propose a novel deep saliency model, FastSal, that is more computationally efficient than the existing state-of-the-art and delivers performance on par with many more complex models. The model uses an efficient MobileNetV2~\cite{sandler2018mobilenetv2} backbone and is trained using a form of distillation using the outputs a more complex DeepGaze~II~\cite{kummerer2016deepgaze} on a large dataset. The proposed model has $>10\times$ fewer parameters than comparable state-of-the-art and is the only one tested that achieves real-time ($>$30~fps) performance on a CPU, making it well-suited to mobile deployment and acceleration of downstream tasks.

\section{Related Work}
Substantial research has been invested into improving the accuracy of saliency prediction models in recent years. A key biological function of attention in the human visual system is to efficiently allocate resources. This preprocessing task requires not only accurate but also fast and efficient models. There have been several important developments in both models of human visual attention and in computationally efficient neural models over the past few years. The following reviews some of the more important developments related to this work.

\subsection{Saliency}
The goal of saliency prediction is to predict which part of images or video attract human attention. Before the emergence of deep neural networks, researchers~\cite{itti1998model} used linear combinations of features that correlated with attention in the human visual system. Since the emergence of deep neural networks, bottom-up approaches that attempt to learn a  computational model to map directly from stimulus images to saliency maps are the most common. In an early work in this paradigm, Pan et al.~\cite{pan2016shallow} proposed a shallow and a deep neural networks for this task. In the shallow net, 3 convolutional layers, 3 max pooling layers and 2 fully connected layers, with 64.4M total parameters, are used to generate an output of $\frac{H}{2} \times \frac{W}{2}$ with input size $H \times W$. The deep network (SalNet) uses a VGG-like structure with 8 convolutional layers and 2 max pooling layers and 1 deconvolutional layer with total of 25.8M parameters. SalNet generates an output that is the same size as its input.
The SalNet authors uses a deconvolutional layer instead of bilinear upsampling to improve performance. The deconvolutional layer can be seen as an early attempt to use a decoder in saliency detection. In state-of-the-art saliency models, the encoder-decoder structure is essentially the de-facto standard design.

The decoder can be broadly defined as any computational approaches that transforms image features (an image representation) to a saliency map. There are variety of decoders that can be used for this task.
SalGAN~\cite{pan2017salgan} use a reverse version of VGG-16 as a decoder. The decoder contains blocks of convolutional layers that are arranged so that the channel dimension of each the five block outputs are 512, 512, 256, 128, and 1. Apart from using binary cross entropy, the authors also propose to use (conditional) discriminator loss to encourage the generated saliency map and generated saliency map to share statistics. 
DeepFix~\cite{kruthiventi2017deepfix} proposes to use an inception-like block to exploit image features by using convolutional kernels of different sizes, channels, and dilations. These blocks contain a convolutional layer with a large dilation to capture information in a larger receptive field. Positional encoding with a Gaussian prior, inspired by the central bias, is also used. 
ML-NET~\cite{cornia2016deep} use concatenated image features extracted from multiple points in a deep neural net instead of just features from the last convolutional layer, which has a very high semantic level. Inclusion of some low-level image representations seems a reasonable approach to increase performance.
DeepGaze II~\cite{kummerer2016deepgaze} also take advantage of features from different levels; instead of 3 layers in ML-Net, 5 layers of features are resized and concatenated. This multi-level representation is then passed to a decoder, called a readout network, which consists of 4 convolutional layers with output sizes of 16, 32, 2, 1. A Gaussian kernel is also convolved with the output to regularize it.
EML-Net~\cite{jia2020eml} emphasizes the encoder rather than the decoder. The authors use 2 deep networks pretrained on 2 different datasets to extract image representations. The decoder contains multiple simple convolutional layers and uses features from multiple encoder layers. EML-Net clearly demonstrates that the complexity of the decoder can be reduced with better image representations.
ML-NET, EML-NET, and DeepGaze II use features from multiple semantic levels and all achieve good performance, which is consistent with the observations that low-level features and high-level semantics both contribute to human attention~\cite{frintrop2010computational}.

\subsection{Fast Networks}
Fast and compact models that generate useful image representations are particularly important in constrained computing environments. There has been progress in building neural networks that can run fast enough on devices with limited compute capacity like mobile phones. 
MobileNetV2~\cite{sandler2018mobilenetv2} uses an inverted bottleneck unit as the basic building block in which pointwise convolutions with batch normalization are used to map between channel representations with different dimensions, and depthwise or separable convolution is used to reduce FLOPs. In a bottleneck structure, the number of  input and output channels needs to be the same and the middle layer has fewer channels than the input and output. The inverted bottleneck, on the other hand, has more channels in the middle layer. ShuffleNet~\cite{zhang2018shufflenet} proposes a basis bottleneck unit that uses pointwise group convolution as a substitute for normal convolution to reduce FLOPS, along with channel shuffle to enable cross group information flow. Depthwise convolution is also used to reduce input channels, as in bottleneck units. 
ShuffleNet V2~\cite{ma2018shufflenet} rearranges the bottleneck structure. Channel split is applied to separate the input of the unit into 2 branches. Concatenation is used instead of addition to fuse information flow of the 2 branches. Channel shuffle is used after the concatenation of the flow. Group convolution is also reverted to standard convolution with batch normalization. The network design shows that group convolution, and its special case depthwise convolution, is the key factor in reducing the theoretical number for float multiply-add operations. 
In EfficientNet~\cite{tan2019efficientnet}, the authors attempt a holistic approach to network design. Since network depth, width, and image resolution are the 3 factors that change the computational complexity of a model, they designed experiments using a compound scaling method to find optimal configurations of depth, width, and resolution. The depth is controlled by stacking ResNet-30 bottleneck units or MobileNetV2 inverted bottleneck units. Similar work has been done to improve the efficiency of object detection. EfficientDet~\cite{tan2020efficientdet} uses a repeated unit similar to the feature pyramid network~\cite{lin2017feature} to encode information from different CNN layers. It also uses skip connections between repeated blocks to improve gradient flow.
\section{Proposed Approach}

\subsection{Lightweight model with hierarchical adaptation}
One hindrance to wide application of saliency detection is the computational cost of inference in modern models. This is especially important for saliency prediction models using neural networks because of the large number of parameters. As noted previously, there are several popular lightweight CNNs proposed for mobile devices such as ShuffleNet~\cite{zhang2018shufflenet, ma2018shufflenet}, MobileNetV2~\cite{sandler2018mobilenetv2}, and EfficientNet~\cite{tan2019efficientnet}. Our experiments revealed that of these networks, MobileNetV2 works particularly well for saliency map generation. We thus choose this network as a backbone to extract image representations.  

A typical saliency prediction architecture includes an encoder and a decoder, similar to the structure used in autoencoders in which an encoder extracts latent features and a decoder translates those features back to images. We see several different design choices in networks designed for saliency prediction: 1) the encoder and decoder are about the same size (e.g. SalGAN); 2) the decoder is smaller than the encoder (e.g. EML-Net~\cite{jia2020eml} and ML-Net~\cite{cornia2016deep}); 3) the decoder is larger than the encoder (e.g. SAM-ResNet~\cite{cornia2018predicting}). The latter structure often uses a recurrent neural network to decode image representations, meaning the decoder is the most computationally complex component. Intuitively, however, the saliency map contains less information than the original image, implying a large decoder may be unnecessary.
Furthermore, it is clear from the state-of-the-art that a larger decoder does not ensure high accuracy.

\begin{figure*}
    \centering
    \includegraphics[width=514px]{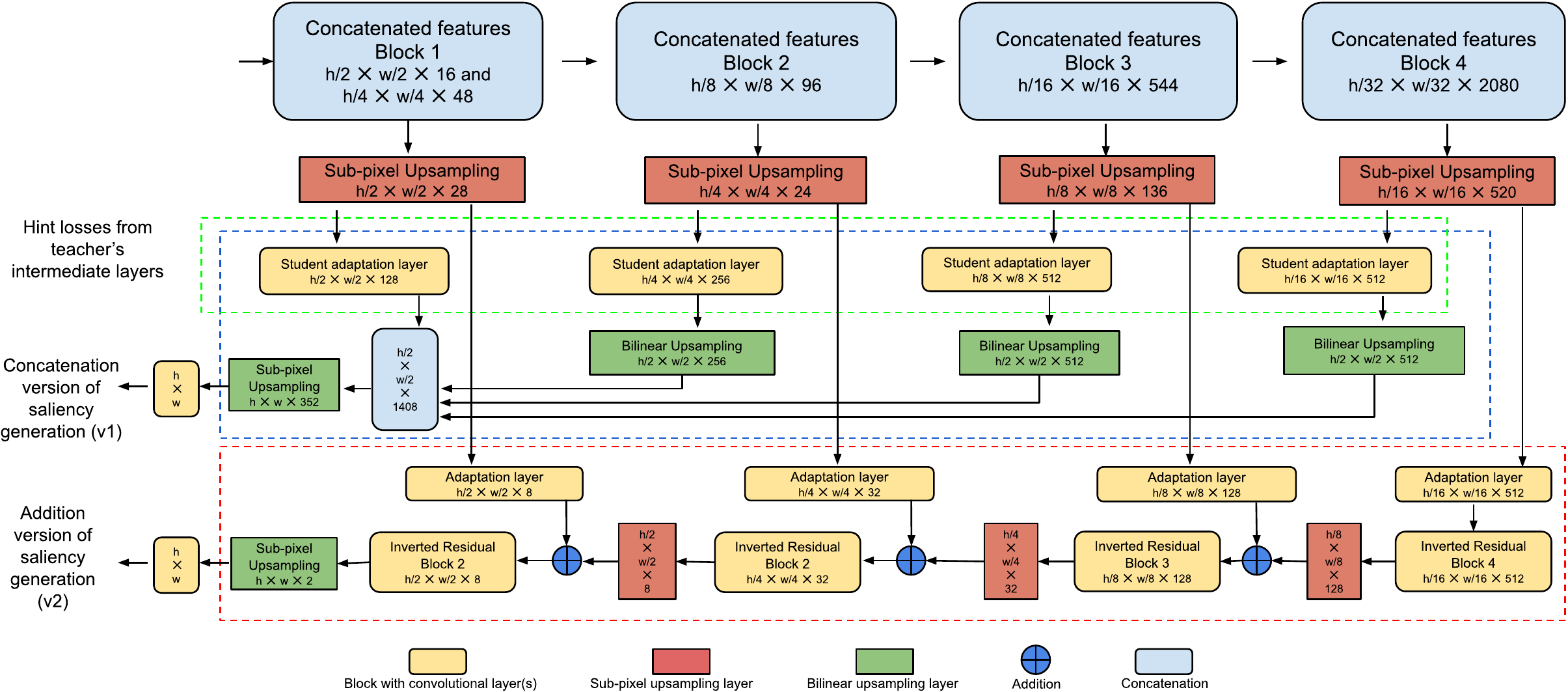}
    \caption{FastSal network structure. The output size is shown at the bottom of each module.}
    \label{fig:fastsal_net}
\end{figure*}

Figure~\ref{fig:fastsal_net} shows the proposed network architecture for efficient saliency map generation. The  principles we follow to design networks for fast saliency prediction include: 1) a lightweight backbone for image feature computation, where lightweight here refers to a small number of weights and fewer floating point operations; 2) a compact decoder designed to exploit features from the backbone network with minimal computational cost. Many researchers have used features from different hierarchical layers~\cite{cornia2016deep, kummerer2016deepgaze, jia2020eml}; we take this further by using the output from all 18 intermediate layers in MobileNetV2. Features with same size are concatenated on the channel axis to form 4 blocks of features that cover low-level to high-level semantics. Obviously after channel-wise concatenation some of blocks have a large number of channels. We avoid this by applying sub-pixel upsampling~\cite{shi2016real}, which rearranges input data with shape $H \times W \times C$ to $2H \times 2W \times \frac{C}{4}$. By using sub-pixel upsampling we reduce the number of channels significantly and increase image size without using convolutional layers, which is exactly what is needed in fast saliency map generation. We developed 2 types of decoder to subsequently merge and transform features from different feature blocks to the saliency map. The key difference in these decoders is the way in which blocks are merged: the first uses concatenation and the second addition. Section~\ref{cat_add_features} provides a detailed description of both types of decoder. 

\subsection{Compact model through knowledge distillation}
There are many techniques in the literature to reduce the size of over-parameterized neural networks such as pruning~\cite{molchanov2016pruning}, matrix factorization~\cite{jaderberg2014speeding}, and knowledge distillation~\cite{hinton2015distilling, mirzadeh2019improved, chen2017learning}. There are some limitations in pruning; for instance, the resulting network usually has similar network structure, depth and input space, which may not be desirable. One advantage of knowledge distillation is that it treats neural networks like ``black boxes.'' To implement knowledge distillation, two networks are needed: one is termed the teacher network and the other the student network. The teacher is normally large and well-trained while the student is usually much smaller and either shallower or narrower.

Buciluǎ et al.~\cite{bucilua2006model} propose an approach to replicate output of an ensemble of models by training a neural network. The work was extended by Ba and Caruana~\cite{ba2014deep} to train compact networks with shallower and wider architectures. Originally knowledge distillation was applied to train compact models in a supervised manner; that is to say during training process both manual labels and teacher supervision are used together. However, the nature of student-teacher network enables weak supervision by letting the student network only use the trained teacher supervision. In the context of saliency prediction, there are many publicly available trained models with strong performance. Those models can be used to pretrain our fast saliency model and/or facilitate generation of final saliency maps. In the paper our fast saliency generation networks are the student networks and pretrained larger SALGAN and DeepGazeII networks are used as the teacher network. To validate the feasibility of this hypothesis, we implemented two experiments. There are two key steps in the experiments: 
1) Generate pseudo saliency maps using trained models. We use SalGAN and DeepGazeII to generate pseudo labels for SALCICON 2017 and COCO dataset in our experiments (see Section~\ref{sec:dataset} for details of the datasets). For the COCO dataset, we generate pseudo saliency maps using both DeepGazeII and SalGAN. For SALCICON, we only use SalGAN as the teacher and pseudo label generator.
2) The backbone network is initialized with pretrained weights from ImageNet. We then pretrain the backbone network with the student adaptation layer, which is $1 \times 1$ convolutional layer whose output size matches the teacher's intermediate layers. The technique is referred as a \textit{hint loss} in knowledge distillation and it improves performance comparing with directly learning knowledge from target generated by teacher~\cite{romero2014fitnets}. Although by applying the adaptation layer, the teacher's intermediate outputs are not required to have same dimensions as those in the student's network, the width and height of the features maps are expected to be the same. By using the student adaption layer we are encouraging the intermediate output from the student network to be close to those of the teacher's network. In our pretraining experiments the hint loss is defined as:
\begin{equation}
    L_{\text{hint}}(\hat{y}, \Bar{y}) = \frac{1}{N}\sum_{l=1}^L \sum_{n=1}^N {(\Bar{y}_{nl} - \hat{y}_{nl})^2},
\end{equation}
where $N$ is the number of elements in the flattened feature maps, $L$ is number of intermediate layers (in our experiment it is 4), 
$\hat{y}$ are the adapted intermediate outputs from the student and $\Bar{y}$ are the intermediate outputs from the teacher network. The hint loss is the mean squared error from each intermediate layer summed over the four intermediate layers.

\subsection{Concatenation and addition of hierarchical features}
\label{cat_add_features}
We developed 2 types of network that decode hierarchical features to produce a saliency map: FastSal (C) a version using feature concatenation, and FastSal (A) a version using feature addition. These versions correspond to the large blue and red dashed boxes in Figure~\ref{fig:fastsal_net}. In FastSal (C), outputs from student adaptation layers are upsampled to the same size of $\frac{H}{2} \times \frac{W}{2}$ and concatenated on the channel dimension to form a feature map of size $\frac{H}{2} \times \frac{W}{2} \times 1408$. Sub-pixel upsampling is then used to resize the feature map to match the input image size, and one additional convolutional layer is used to reduce the features to a single channel.

\begin{figure}
    \centering
    \includegraphics[width=160px]{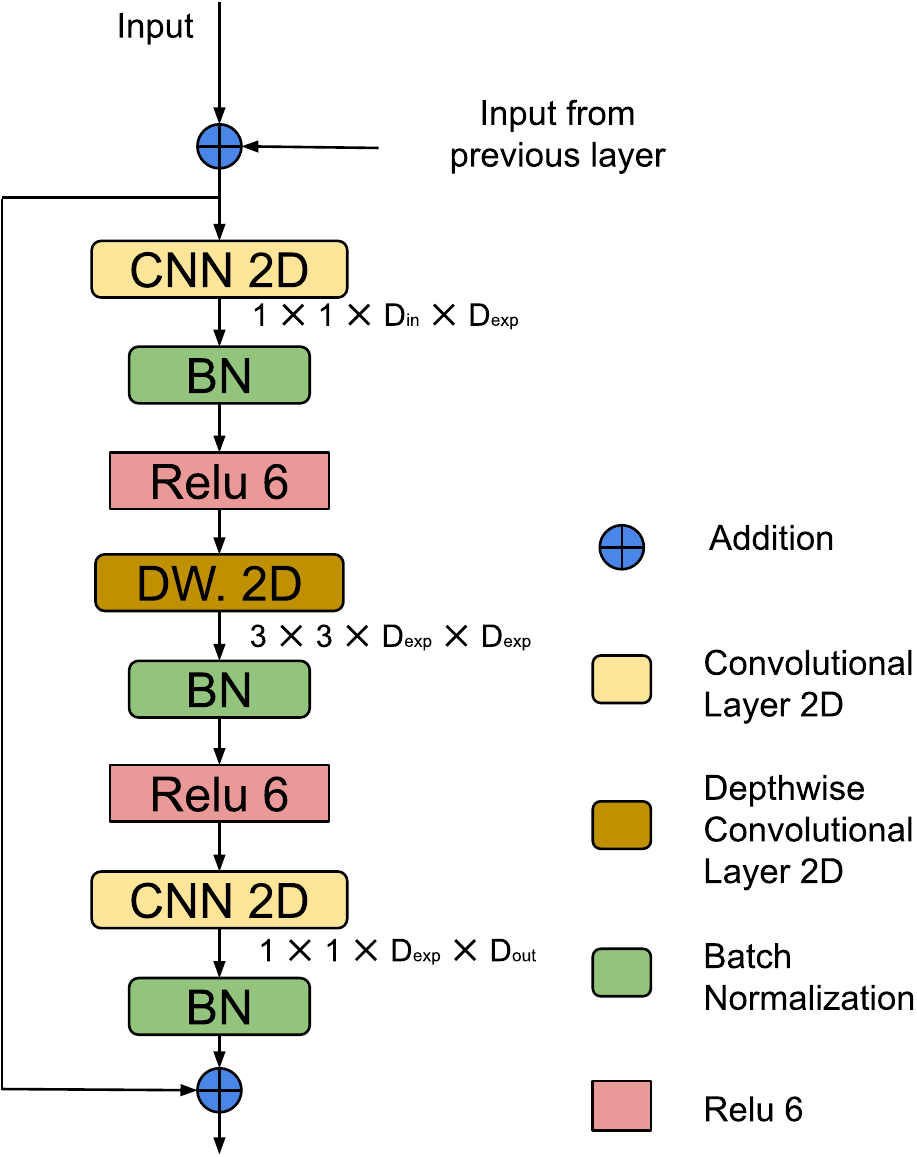}
    \caption{The structure of modified Inverted Residual Block. The size of each convolution kernel is marked below each convolutional block. $D_{exp}$ is the dimension of a hidden separable convolutional layer $D_{exp} = 2 D_{in}$.}
    \label{fig:inverted_block}
\end{figure}

FastSal (A) is inspired by how the FPN~\cite{lin2017feature} handles features from feature pyramids generated from different semantic levels. Examining the structure of the concatenation version of decoder revealed that a substantial fraction of the floating point operations are used in the student adaptation layer to match the channel count of the teacher network. As a solution, we substitute the student adaptation layer with an adaptation layer that reduces the channel count using a more radical approach. The output of the adaptation layer is then passed to a inverted residual block, which similar to that used in MobileNetV2, but it adds features from its layer and resized features from the previous layer before feeding the result into the original MobileNetV2 inverted residual block. Figure~\ref{fig:inverted_block} shows the internal structure of a modified inverted residual block. We use an expansion rate of 2 in our inverted residual block.

We use a different loss function for training the SalGAN and DeepGaze II teachers. SalGAN generates a traditional saliency map with values quantized to range from 0 to 255. DeepGaze II outputs a probability distribution, which means our network has to generate a similar output to learn from this teacher. The per-sample loss we use for the SalGAN teacher is:
\begin{equation}
\label{eq_salgan_loss}
    \begin{split}
        L = -\frac{1}{N}\sum_{i,j}{y_{ij}\log \sigma(\hat{y}_{ij}) + (1-y_{i,j})\log(1-\sigma(\hat{y}_{ij})) } \\
        -\frac{1}{N}\sum_{i,j}{\Bar{y}_{ij}\log \sigma(\hat{y}_{ij}) + (1-\Bar{y}_{ij})\log(1-\sigma(\hat{y}_{ij})), }
    \end{split}
\end{equation}
where $N$ is the number of pixels, $\sigma(\hat{y}_{ij})$ is the predicted saliency of pixel $i,j$, $y$ is the ground truth saliency map, and $\Bar{y}$ is the generated pseudo saliency from the pretrained SalGAN teacher. The predicted saliency is activated with a sigmoid $\sigma$ so that each pixel can be treated as a Bernoulli random variable. This allows the use of the total binary cross-entropy loss between the prediction and the manual labels and teacher's supervision. 

The per-sample loss function used with the DeepGaze II teacher is:
\begin{equation}
\label{eq_deepgaze_loss}
    \begin{split}
        L &= \sum_{i,j}\Bar{y}_{i,j}\log\frac{\Bar{y}_{i,j}}{g(\hat{y}_{i,j})} +  1 - \frac{\sum_{i,j}\Bar{y}_{i,j} g(y_{i,j})}{\sqrt{\sum_{i,j}\Bar{y}_{i,j}^2}\sqrt{\sum_{i,j} g(y_{i,j})^2}}\\
          &-\frac{1}{N}\sum_{i,j}{f(\Bar{y}_{i,j})\log \sigma(\hat{y}_{i,j})  + (1-f(\Bar{y}_{i,j}))\log(1-\sigma(\hat{y}_{i,j})), }
    \end{split}
\end{equation}
where $\hat{y}$ is the unnormalized output from neural network, and $\Bar{y}$ is the generated probability distribution from the DeepGaze II teacher. $g$ is an activation function that converts the output of a neural network to probability distribution; we use the softmax here. $f$ converts a saliency map represented as a probability distribution (i.e. sums to one) to a traditional saliency maps with pixels ranging from 0 to 1. We use min-max normalization which maps the smallest value to 0 and the largest to 1 with values in between scaled linearly. The loss function consists of 3 parts. The first is a KL-divergence term that measures the difference between teacher and student's predicted distribution. The second term is cosine similarity term that measures the similarity of the predictions and pseudo-labels. Unlike KL-divergence this metric is symmetric. We subtract the cosine similarity from 1 to make it a loss and ensure the result is positive. The third term is the binary cross entropy loss we also use for the conventional saliency map, which treats each pixel in the saliency map as if it were an independent Bernoulli random variable.

\section{Experiments}

\subsection{Datasets}
\label{sec:dataset}
\textbf{MIT300}~\cite{judd2012benchmark} includes 300 natural indoor and outdoor scenes. The maximum dimension of the images are 1024 pixels. To collect the eye fixations 39 participants aged 18-50 years old are required to wear a eye-tracking device when they are given stimulus images. The participants are free to view the images for 3 seconds. The MIT300 data set is used for testing purposes only: the ground truth is kept private to avoid overfitting. The MIT1003 dataset~\cite{judd2009learning} is similar but contains more images and the ground truth is publicly available. Some researchers use this dataset to test models before submitting to MIT300.

\textbf{SALICON}~\cite{jiang2015salicon} is a dataset that is part of the Large-Scale Scene Understanding (LSUN) challenge. The challenge was held twice in 2015 and 2017, so 2 datasets were available, namely SALICON 2015 and 2017. The dataset is also focused on understanding saliency in natural scene images. The dataset is much larger than MIT300 and MIT1003, with 10K training, 5K validation, and 5K (unreleased) test samples. The images in SALICON are a subset of the Microsoft COCO dataset~\cite{lin2014microsoft}. Due to the large number of images, it would be expensive to collect eye-fixations from multiple participants for each image. Fortunately, researchers have noticed a correlation between eye-fixations and mouse trajectories when subjects are viewing artificially foveated imagery~\cite{jiang2015salicon}. The mouse trajectories can be used as proxy visual saliency and models have been developed to transfer mouse trajectories to eye fixation maps. The ground-truth SALICON annotations contains fixations generated from mouse trajectories. Many researchers have used this dataset for pretraining saliency models and reported good performance~\cite{kummerer2016deepgaze}.

\textbf{COCO} is large-scale dataset used for object detection, segmentation, and captioning. It contains pixel-wise annotation of natural scenes. The dataset contains 330K images with more than 200K labeled, 1.5 million object instances, 80 object categories, 91 stuff categories, 5 captions per image, 250k people with keypoints. Since we are experimenting the validity of using well-trained teacher to transfer knowledge to a student via generating pseudo labels for the student, sufficient training samples are important for the student to learn good representations. We use images from the 2019 COCO detection task to generate large-scale pseudo saliency map dataset. This sub task of COCO contains $\approx 118$K training images, 40K test images, and 5K validation images. We use all of these images for pseudo label generation.

\subsection{Implementation}
\label{sec:resuts}
The FastSal networks are implemented in PyTorch. Experiments were run on a NVIDIA GTX 1080 Ti. To evaluate the speed on CPU an Intel-i7-6850K is used. During pretraining the network is trained with SGD for 100 epochs using the hint loss. The initial learning rate is set to 0.01 and the learning rate is decayed 10 times at epoch 15, 30, and 60. Fine tuning the SALGAN and DeepGaze II teacher uses the losses in equation \ref{eq_salgan_loss} and \ref{eq_deepgaze_loss} respectively. The same learning rate and optimizer is used as in pretraining. Input image sizes for pretraining and fine tuning are $192\times256$.

\subsection{Comparison with the state-of-the-art}

\begin{table*}[ht]
\centering
\caption{Performance comparison on MIT300 (FastSal concatenation version). $^1$ and $^2$ are fine tuned with Equation \ref{eq_salgan_loss} and \ref{eq_deepgaze_loss}.}
\begin{tabular}{llllllll}
\toprule
 & AUC$\uparrow$ & sAUC$\uparrow$ & NSS$\uparrow$ & CC$\uparrow$ & KLDiv$\downarrow$ & SIM$\uparrow$ & IG$\uparrow$\\ \midrule
EML-NET~\cite{jia2020eml} & 0.8762 & 0.7469 & 2.4876 & 0.7893 & 0.8439 & 0.6756 & N/A \\ 
DeepGaze II~\cite{kummerer2016deepgaze}  & 0.8733 & 0.7759 & 2.3371 & 0.7703	& 0.4239 & 0.6636 & 0.9247 \\ 
GazeGAN~\cite{che2019gaze} & 0.8607 & 0.7316 & 2.2118 & 0.7579 & 1.3390 & 0.6491  & N/A \\ 
CASnet-II~\cite{fan2018emotional} & 0.8552 & 0.7398 & 1.9859	& 0.7054 & 0.5857 & 0.5806 & N/A \\ 
SAM-ResNet~\cite{cornia2018predicting} & 0.8526	& 0.7396 & 2.0628 & 0.6897 & 1.1710	& 0.6122 & N/A \\ 
SalGAN~\cite{pan2017salgan} & 0.8498	& 0.7354 & 1.8620 & 0.6740 & 0.7574	& 0.5932 & N/A \\ 
SAM-VGG~\cite{cornia2018predicting} & 0.8473 & 0.7305 & 1.9552 & 0.6630	& 1.2746 & 0.5986 & N/A\\ 
DVA~\cite{wang2017deep} & 0.8430 & 0.7257 & 1.9305 & 0.6631	& 0.6293 & 0.5848 & N/A \\ 
ML-Net~\cite{cornia2016deep} & 0.8386	& 0.7399 & 1.9748 & 0.6633 & 0.8006	& 0.5819 & N/A\\  \midrule
FastSal (C)$^1$ & 0.8635 & 0.7261 & 2.1158 & 0.7448 & 0.7086 & 0.6422 & N/A \\ 
FastSal (C)$^2$ & 0.8684 & 0.7701 & 2.1913 & 0.7507 & 0.4665 & 0.6456 & 0.8355 \\ \bottomrule
\end{tabular}
\label{tab:mit300}
\end{table*}

\begin{table*}[ht]
\centering
\caption{Performance comparison on the SALICON 2017 test set.}
\begin{tabular}{llllllll}
\toprule
 & AUC$\uparrow$ & sAUC$\uparrow$ & NSS$\uparrow$ & CC$\uparrow$ & KLDiv$\downarrow$ & SIM$\uparrow$ & IG$\uparrow$\\ \midrule
EML-NET~\cite{jia2020eml} & 0.866 & 0.746 & 2.050 & 0.886 & 0.520 & 0.780 & 0.736 \\  
SAM-ResNet~\cite{cornia2018predicting} & 0.865 & 0.741 & 1.990 & 0.899 & 0.610 & 0.793 & 0.538 \\ 
SalGAN~\cite{pan2017salgan} & 0.864 & 0.732 & 1.861 & 0.880 & 0.288 & 0.772 & 0.775 \\  
GazeGAN~\cite{che2019gaze} & 0.864 & 0.736 & 1.899 & 0.879 & 0.376 & 0.773 & 0.720 \\  \midrule
FastSal (C) & 0.863 & 0.732 & 1.845 & 0.874 & 0.288 & 0.768 & 0.770\\ 
FastSal (A) & 0.862 & 0.731 & 1.828 & 0.870 & 0.291 & 0.764 & 0.760\\ \bottomrule
\end{tabular}

\label{tab:salicon2017}
\end{table*}

\begin{figure*}
    \centering
    \includegraphics[width=514px]{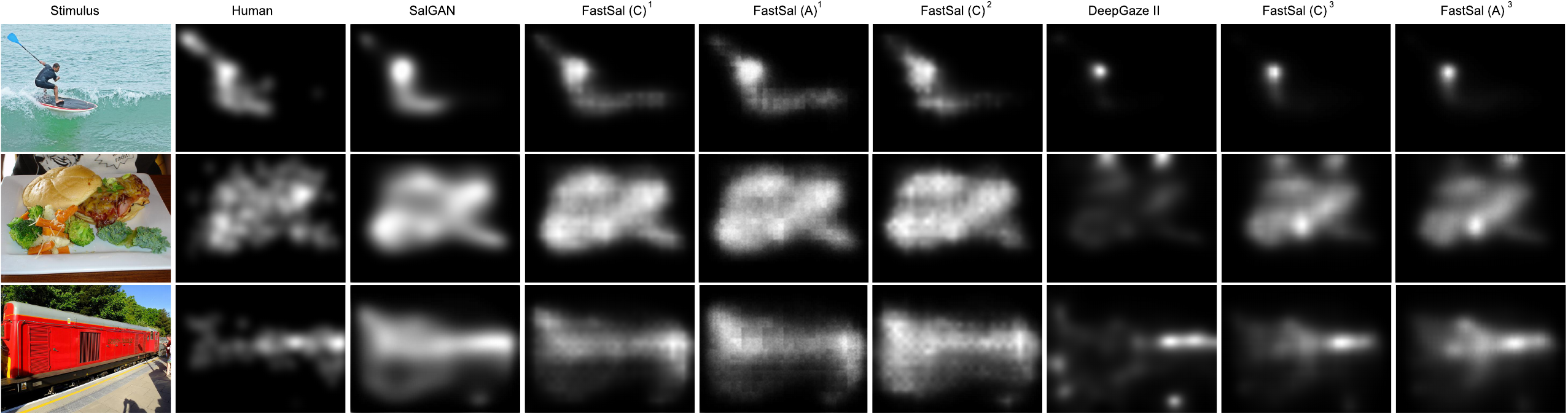}
    \caption{Qualitative comparison among ground truth and various saliency models. $^1$ is pretrained and fine-tuned on SALICON 2017 with SalGAN and the ground truth. $^2$ is trained with ground truth only. $^3$ is pretrained and fine-tuned on COCO detection 2019 with DeepGaze II and the ground truth.}
    \label{fig:examples}
\end{figure*}

Table~\ref{tab:mit300} and \ref{tab:salicon2017} show the overall performance achieved on the MIT300 and SALICON 2017 test set. The proposed model achieves top 3 performance in terms of the AUC metric on MIT300. It is among the best models on SALICON 2017, especially with regards to KL-divergence and information gain. We used teacher supervision from DeepGaze II on MIT300, and supervision from a SalGAN teacher on SALICON 2017. With this student-teacher setup we find that the student is able to successfully imitate the output of the teacher network and find that the performance of FastSal is similar to the teachers' performance in both datasets. Figure~\ref{fig:examples} shows a qualitative comparison of student and teacher for several examples.

\subsection{Speed and model size}
FastSal was designed to achieve competitive performance with less computational cost. We use the number of parameters to evaluate model size and GFLOPs to evaluate the theoretical speed. Table~\ref{tab:flops_size} shows the GFLOPs and parameter counts (millions) of various state-of-art models.
\begin{table}[ht]
\centering
\caption{Estimated GFLOPs and model size (millions of parameters) for the proposed and several other state-of-art models. GFLOPs are evaluated using an image input size of $192\times256$}
\begin{tabularx}{\columnwidth}{Xrrc}
\toprule
       &GFLOPs & Size (M) & Output size \\ \midrule
SalGAN~\cite{pan2017salgan} &91.46 & 31.92 & $192\times256$     \\  
DeepGaze II~\cite{kummerer2016deepgaze} &20.22 & 20.44 & $192\times256$     \\  
EML-Net~\cite{jia2020eml} &16.24 & 47.09 & $192\times256$     \\  
CASNet-II~\cite{fan2018emotional} &37.62 & 42.01 & $6\times 8$     \\  
SAM-ResNet~\cite{cornia2018predicting} & 72.9 & 70.04 & $192\times256$     \\  
SAM-VGG~\cite{cornia2018predicting} & 59.4 & 51.83 & $192\times256$     \\  
ML-Net~\cite{cornia2016deep} & 39.42 & 15.45 & $24\times32$     \\  
DeepFix~\cite{kruthiventi2017deepfix} & 59.82 & 28.73 & $24\times32$ \\  \midrule
FastSal (C) & \textbf{1.32} & \textbf{2.57} & $192\times256$ \\ 
FastSal (A) & \textbf{1.32} & 3.65 & $192\times256$ \\ \bottomrule
\end{tabularx}
\label{tab:flops_size}
\end{table}

\begin{figure}
    \centering
    \includegraphics[width=1.0\columnwidth]{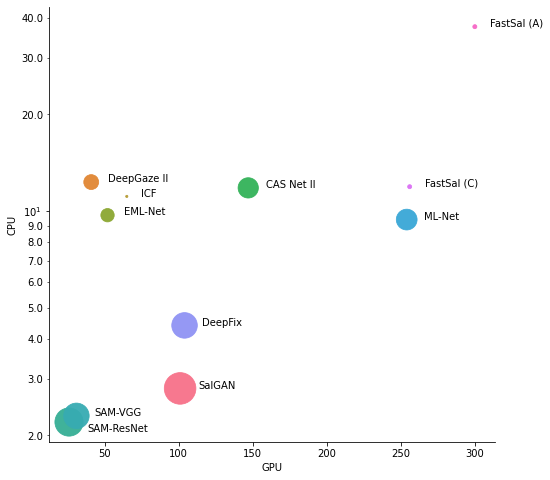}
    \caption{Frames per second on GPU and CPU of various saliency models. The y-axis is on a log scale. Larger dots represent higher FLOPs.}
    \label{fig:cpu_gpu}
\end{figure}

In addition to estimating the theoretical FLOPs of each model, we also benchmark the actual speed on GPU and CPU hardware to compare the speed of state-of-art models with our FastSal in reality. The benchmark uses input images of size $192 \times 256$ and 100 iterations. Frame per second is computed using the mean processing time for each image. Figure~\ref{fig:cpu_gpu} shows the benchmark results for various saliency models. If a model is located at the right side of the figure, it means the model has high FPS on the  GPU (GTX 1080 Ti); if it is at top it indicates that the model has high FPS on a CPU. Both FastSal (C) and FastSal (A) appear in the top-right of the figure indicating fast processing speed on real hardware. FastSal (A) in particular reached $\approx 300$ FPS on the GPU and more than 35 FPS on the CPU. This speed could be further improved in future: the depthwise convolutional layer is still relatively new and is not as well-optimized as the conventional convolutional layer at present. We also tested different implementations of depthwise convolutional layers, finding the TensorFlow implementation is about 2-3 times faster than PyTorch.

\subsection{Backbone variants}
Table~\ref{tab:effnet} compares the performance of two saliency networks that use EfficientNet (b0) and MobileNetV2 as the encoder and a 6-layer CNN as the decoder. The intermediate output dimensions of the decoder are 512, 512, 256, 128, 64, 1 to match that of SalGAN. A hint loss is computed between student and SalGAN teacher. MobileNetV2 has better performance for all metrics and lower hint loss.
\begin{table}[t]
\caption{Results on SALICON 2017 (validation) with different backbones.}
\centering
\begin{tabular}{llllll}
\toprule
       &sAUC$\uparrow$ & bAUC$\uparrow$ & NSS$\uparrow$ & CC$\uparrow$ & hint loss$\downarrow$\\ \midrule
EfficientNet (b0) & 0.7275 & 0.8421 & 1.7623 & 0.8579 & 0.0617 \\ 
MobileNetV2 & \textbf{0.7365} & \textbf{0.8450} & \textbf{1.8163} & \textbf{0.8751} & \textbf{0.0526} \\ \bottomrule 
\end{tabular}
\label{tab:effnet}
\end{table}

\subsection{Ablative studies}
We also tested the effectiveness of using knowledge distillation in training the networks. Table~\ref{tab:teacher_student} reports the results. Here we focus on NSS and CC metrics as suggested in~\cite{bylinskii2018different}, and test combinations of different pretraining and fine-tuning approaches. SALICON 2017 is used for both pretraining and fine-tuning. The first column in the table indicates whether the network is pre-trained using the hint loss. The second and third columns concern the fine tuning procedure: the second indicates if SalGAN is used to generate pseudo saliency maps and the third indicates whether the ground truth is used. The results show that the combination of all three gives the best performance.

\begin{table}[t]
\centering
\caption{Ablation study on distillation with SalGAN as the teacher. Results are reported on the SALICON 2017 validation set and MIT1003 using FastSal (C).}
\begin{tabular}{lllllll}
\toprule
\multirow{2}{*}{Pretrain} & \multirow{2}{*}{Finetune} & \multirow{2}{*}{GT} & \multicolumn{2}{l}{SALICON 2017} & \multicolumn{2}{l}{MIT 1003} \\ \cmidrule{4-7} 
 &  &  & NSS$\uparrow$ & CC$\uparrow$ & NSS$\uparrow$ & CC$\uparrow$ \\ \midrule
\xmark & \cmark & \xmark & 1.7749 & 0.8603 & 1.9133 & 0.6133 \\
\cmark & \cmark & \xmark & 1.7789 & 0.8604 & 1.9192 & 0.6141 \\
\xmark & \xmark & \cmark & 1.7861 & 0.8604 & 1.9106 & 0.6088 \\
\xmark & \cmark & \cmark & 1.8004 & 0.8697 & 1.9412 & 0.6202 \\
\cmark & \cmark & \cmark & \textbf{1.8129} & \textbf{0.8717} & \textbf{1.9503} & \textbf{0.6213} \\
\bottomrule
\end{tabular}
\label{tab:teacher_student}
\end{table}


\subsection{Performance on downstream tasks}
We apply FastSal to a downstream task to establish if the model gives comparable results. The Saliency Bag of Words (SalBoW) approach~\cite{mohedano2018saliency} uses saliency maps to generate better image representations for instance retrieval. It is based on weighting different local features with saliency to improve global image representations. Table \ref{tab:salbow} gives mAP results using FastSal in place of SalGAN for various image retrieval tasks. With FastSal pretrained and fine tuned using SALICON and SalGAN, we achieve slightly improved mAP than the SalGAN network used in the original paper.
\begin{table}[t]
\caption{Results (mAP) of image retrieval tasks using SALBOW on the INSTRE, Oxford, and Paris datasets. +qe indicates average query expansion $^1$ is fine-tuned on SALICON with SalGAN. $^2$ is fine-tuned on COCO with DeepGaze II and sigmoid activation output. $^3$ is fine-tuned on COCO with DeepGaze II and softmax  output.}
\centering
\begin{tabular}{lllllll}
\toprule
      &INSTRE & +qe & Oxford & +qe & Paris &+qe \\ \midrule
SalGAN &0.698 &0.757 &0.746 &0.778 &0.812 &0.830 \\ 
FastSal (C)$^1$ &\textbf{0.703} &\textbf{0.765} &0.751 &0.794 &0.818 &0.832 \\ 
FastSal (C)$^2$ &0.693 &0.751 &\textbf{0.756} &\textbf{0.798} &\textbf{0.821} &\textbf{0.836} \\ 
FastSal (C)$^3$ &0.681 &0.736 &0.747 &0.787 &0.813 &0.826 \\ \bottomrule
\end{tabular}
\label{tab:salbow}
\end{table}

\section{Conclusion}

This paper proposed FastSal, a new fast saliency model suitable for inference on constrained computing devices. The proposed model is significantly smaller than other state-of-the-art with only $2.57\times 10^6$ parameters, which amounts to less than 10MB uncompressed single precision floating point memory. The computational complexity of the model, approx $1.32\times 10^6$ FLOPs for a $192\times 256$ image, is orders-of-magnitude lower than comparable state-of-the-art (e.g. $45\times$ lower than DeepFix) while remaining competitive with top models on most metrics. Tests on a downstream task show that the model can be used in place of more complex models like SalGAN without deteriorating performance.

\section*{Acknowledgment}

This publication has emanated from research conducted with the financial support of Science Foundation Ireland (SFI) under grant number SFI/15/SIRG/3283 and SFI/12/RC/2289\_P2.

\bibliographystyle{IEEEtran}
\bibliography{refs}

\end{document}